\documentclass{article}
\usepackage[utf8]{inputenc}
\usepackage{titlesec}
\usepackage{proof}
\usepackage{parskip}
\usepackage{graphicx}
\usepackage{textcomp}
\usepackage{amsfonts}
\usepackage{amsmath}
\usepackage{float}
\usepackage{enumitem}

\title{Learning to Reason \\ \large Theorem proving at first order via reinforcement learning}
\author{Brian Groenke \\ groenke.5@osu.edu}
\date{April 2018}

\begin{document}

\maketitle

\section{Introduction}

Automated theorem proving has long been a key task of artificial intelligence. Proofs form the bedrock of rigorous scientific inquiry. Many tools for both partially and fully automating their derivations have been developed over the last half a century. Some examples of state-of-the-art provers are E (Schulz, 2013), VAMPIRE (Kovács \& Voronkov, 2013), and Prover9 (McCune, 2005-2010). Newer theorem provers, such as E, use superposition calculus in place of more traditional resolution and tableau based methods. There have also been a number of past attempts to apply machine learning methods to guiding proof search. Suttner \& Ertel proposed a multilayer-perceptron based method using hand-engineered features as far back as 1990; Urban et al (2011) apply machine learning to tableau calculus; and Loos et al (2017) recently proposed a method for guiding the E theorem prover using deep nerual networks. All of this prior work, however, has one common limitation: they all rely on the axioms of \textit{classical} first-order logic.
\flushleft
Very little attention has been paid to automated theorem proving for non-classical logics. One of the only recent examples is McLaughlin \& Pfenning (2008) who applied the polarized inverse method to intuitionistic propositional logic. The literature is otherwise mostly silent. This is truly unfortunate, as there are many reasons to desire non-classical proofs over classical. Constructive/intuitionistic proofs should be of particular interest to computer scientists thanks to the well-known Curry-Howard correspondence (Howard, 1980) which tells us that all terminating programs correspond to a proof in intuitionistic logic and vice versa.
\flushleft
This work explores using Q-learning (Watkins, 1989) to inform proof search for a specific system called non-classical logic called Core Logic (Tennant, 2017).
\section{Background}
\subsection{Core Logic}

Core Logic is a system of proof that is both \textit{constructive} and \textit{relevant}. Constructive logics demand that, in order to prove a conjecture, one must not only show its validity but also \textit{demonstrate} a way to "build" it from some set of axioms. More concretely, constructive logics omit several of the familiar rules of inference from classical logic, including (but not limited to) the law of excluded middle (1), double negation elimination (2), and classical reductio (3).
\begin{equation}
\infer{\phi \lor \neg \phi}{}
\end{equation}

\begin{equation}
\infer{\phi}{\neg \neg \phi}
\end{equation}

\begin{equation}
\infer{\phi}{\begin{array}{c} \neg \phi \\ \vdots \\ \bot \end{array}}
\end{equation}

The objection that the constructivist raises here is that such rules allow statements to be proved via indirection (e.g. proof by contradiction) rather than requiring us to explicitly demonstrate its entailment from a set of prior assumptions. Such constraints can be very useful in the context of computation; in formulating instructions for telling a machine how to perform a computation, we do not have the luxury of relying upon the vague and indirect proof methods of classical logic. Here, our logic must be constructive.\textsuperscript{*}

A \textit{relevant} logic is one that further constrains the reasoner to producing proofs in which the conclusion of each inference step satisfies some standard of "relevance" to the premises. There are a number of ways to define such a standard; one naive criterion might be to insist that the conclusion and the premises share atomic formulae. Another is to enforce that assumptions are \textit{discharged}, i.e. actually used, in the resulting proof. One of the most important goals, however, of a relevance logic is to banish the principle of \textit{explosion} (4), a.k.a \textit{ex falso quod libet}, from our set of inference rules.

\textsuperscript{*}{\small It should be noted that Turing completeness requires the introduction of classical assumptions (Howard, 1980); because of this, however, it also brings along the prospect for non-terminating programs. Whether or not such a trade-off is worthwhile largely depends on context, and such a discussion lies outside the scope of this work.}

\begin{equation}
\infer{\phi}{\bot}
\end{equation}

Core Logic achieves this by defining its primitive inference rules with judicious discharge requirements, and more importantly, by enforcing that the major premise for elimination rules \textit{stands proud}, i.e. has no proof work above it. As an example, here is the rule for conditional elimination (more commonly known as \textit{modus ponens}) in Core (5).

\begin{equation}
\begin{array}{c c}
    \infer{\gamma}{
    \begin{array}[b]{c c c}
        \phi \rightarrow \psi &
        \begin{array}[b]{c}
            \vdots \\ \phi
        \end{array} &
        \begin{array}[b]{c c}
            \infer{\psi}{} & \raisebox{0.6em}{\hspace{-0.8em}\scalebox{0.8}{(i)}}\\ \vdots & \\ \gamma &
        \end{array}
    \end{array}}
    & \raisebox{0.4em}{\hspace{-0.65em}\scalebox{0.9}{(i)}}
\end{array}
\end{equation}

For a full discussion of the inference rules of Core Logic, the reader should consult Tennant, 2017.

\subsection{AutoMoL}

\textit{Automated Monadic Logic}, or \textit{AutoMoL} (Groenke, 2017-2018), is a fully automated proof engine for Core Logic. It uses a variant of the backwards chaining algorithm to find possible derivations of a sentence given some set of assumptions. The inference engine is capable of working with any set of well defined inference rules, thus enabling it to find proofs in any variant of Core (including Classical Core, i.e. an extension of Core Logic to include classical rules). The \textit{AutoMoL} proof engine uses an interchangeable \textit{strategy} to determine which actions to apply and in what order. This strategy is the \textit{sine qua non} of the search algorithm. The difference between a good strategy and a bad strategy could be millions of wasted steps investigating fruitless areas of the search space.

\textit{AutoMoL} already provides a built-in search strategy that implements several of the heuristics detailed in \textit{Autologic} (Tennant, 1992); this includes \textit{ordering of choices}, \textit{atomic accessibility}, and the \textit{basic relevance filter}. This heuristics aided strategy is used as the baseline for comparison in this work.

\section{Problems as Graphs}

Wang and Mingzhe, et al (2017), propose a novel way of representing logical formulae as graphs, invariant to variable renaming. They use this to produce graph embeddings from which deep networks can extract syntactic and semantic information. We can apply this idea to guided proof search by representing not only individual sentences as graphs but also the problem as a whole. This is done by first building graph representations for the conclusion as well as each sentence in the set of available assumptions. These graphs are then merged together, with the root nodes of the conclusion and each assumption linked by a single root node for the problem graph. The result is a directed, acyclic, multi-graph. See \textit{Figure 1} for an example.

\begin{figure}
    \centering
    \includegraphics[scale=0.4]{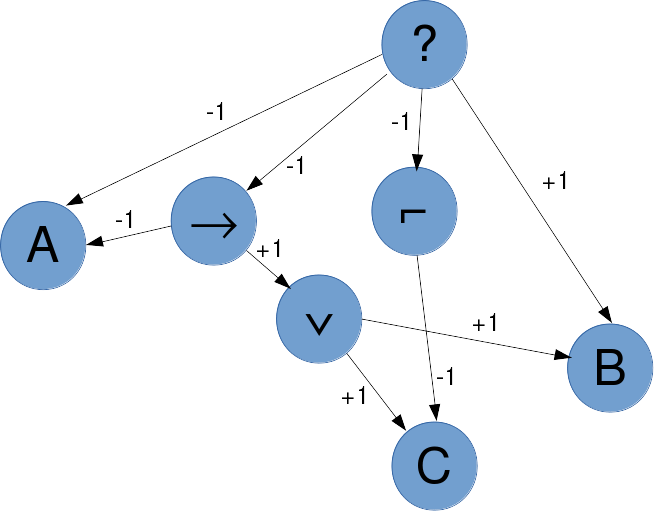}
    \caption{Problem graph for $[A, A\rightarrow (B \lor C), \neg C]\stackrel{?}{\vdash}B$}
    \label{fig:my_label}
\end{figure}

The edge labels of $+1$ or $-1$ encode valuable information about the \textit{parity} of a sub-formula with respect to a parent node. The idea of parity is borrowed from proof theory. In brief, a sub-formula is \textit{negative} if it is the antecedent of a conditional or the complement of a negation, otherwise it is \textit{positive}. We can compute the parity of a sub-formula in our problem graph with respect to any of its ancestors as follows:
\begin{enumerate}
    \item Start at the ancestor node and set our parity "counter" to the parity of its incoming edges. If the node has multiple incoming edges with different parities, we say that its parity is \textit{mixed}. In this case, we can terminate right away as all sub-formulae of any formula with mixed parity must also have mixed parity.
    \item Recurse downward, repeating for all children of the current node for which our sub-formula occurs. If the outward edge to the child node has a negative sign, invert the parity counter.
    \item Halt in each branch once we reach the root node of the desired sub-formula. If one or more branches have differing parity counters, we say the sub-formula is mixed. Otherwise, return the value of the parity counter shared by all branches (either positive or negative).
\end{enumerate}

For a more comprehensive discussion of parity and its implications in proof theory, see Tennant 1992 and 2015.

\section{Q Model for Proof Search}

\subsection{Q Learning}

This project attempts to apply Watkins Q-learning to proof search. Traditional, tabular Q-learning "learns" the expected Q-values (i.e. expected reward) for every seen state-action pair. Q-values are updated using the rule given in (6).

\begin{equation}
    Q(s_t,a_t) \leftarrow (1-\alpha) Q(s_t,a_t) + \alpha (r_t + \gamma \max_{a}{Q(s_{t+1},a)})
\end{equation}

where $\alpha$ is the learning rate, usually between $0$ and $1$, and $\gamma$ is a discounting factor for future rewards.

There is, however, a critical problem with this approach. For contexts where the number of states is very large (possibly even infinite!), such tabular methods are not feasible. Not only would the table be of intractable size, but it would be uselessly sparse; i.e. learned values for "similar" states would be spread across many different cells.

To solve this problem, we can \textit{approximate} the expected Q value as a function of the state/action pairs. Such adaptations of Q-learning have had much recent success in applying reinforcement learning to competitive games (Silver, et al, 2016).

The Q function is approximated as a linear combination of hand-engineered state/action features:

\begin{equation}
    Q(s,a) = w_0 + w_1 f_1(s,a) + w_2 f_2(s,a) + \cdots
\end{equation}

States are represented using the graph detailed in section 3. Actions are represented as an ordered pair $(r,s)$ where $r$ is an available rule of inference, and $s$ is a major premise to use with that rule, if applicable ($s$ will be empty when rules that do not accept a major premise are applied, like introduction rules). Each time an action $a$ is applied to some problem state $s$, the weights for the linear model are updated via gradient descent (8).

\begin{equation}
    w_i \leftarrow w_i + \alpha [r + \gamma \max_{a}{(Q(s_{t+1},a))} - Q(s,a)] f_i(s,a)
\end{equation}

\subsection{Metrics}

Defining metrics for proof search performance is tricky. The two most straightforward measures are the number of inference steps $p$ in the resulting proof and the total number of steps taken (i.e. sub-problems generated) $T$ before completing the proof, where the desire is to minimize both $p$ and $T$. These provide good measures of efficiency between strategies as long as the problem(s) being compared are identical. They cannot be used across different problem sets due to unavoidable variance in the number of steps required (even optimally) to solve each problem. For the purposes of this project, comparing values of $p$ and $T$ across compatible contexts is sufficient.

\subsection{Building a proof search strategy with Q Learning}

Each time the proof engine applies an action, it creates a new \textit{sub-problem}. This sub-problem becomes the new state; it will always have a either a different conclusion or different set of assumptions (or both), and thus must be represented as a new graph. The proof engine then consults the current strategy to figure out the best sequence of actions to apply for the current problem state.

The Q-learning strategy implemented for this project computes Q values for each possible action and then orders the actions by their Q values in descending order, thereby giving preference to actions which are most likely to yield the highest reward, given the current model parameters and problem state. The Q strategy uses the well known $\epsilon$-greedy policy when selecting the action sequence; i.e. with probability $\epsilon$ (where $\epsilon$ is a hyperparameter), it will return a random ordering instead of the optimal one. $\epsilon$ is decayed over time at a fixed rate $\nu$, which is also specified as a hyperparameter.

The reward function (9) is based on the metrics $p$ and $T$ defined in section 4.2. Because the goal is to minimize both $p$ and $T$, the reward function must decrease monotonically with an increase in either value. The use of $\log$ helps mitigate numeric underflow for problems of high complexity.

\begin{equation}
    R(p,T) = \frac{1}{\log{(1 + p)}\log{(1 + T)}}
\end{equation}

\subsection{Designing features}

This section gives a brief summary of a select number of features implemented for this project.

\subsubsection{Rule ordering and basic rule filter}

These features are essentially direct translations of heuristics implemented in baseline search strategy. They are required in order for the Q model to match baseline performance, thereby making the evaluation of new features in the Q model more informative.

The \textit{rule ordering} feature returns a score between $0$ and $1$ indicating preference for the rule given the current problem state. The ordering of the scores are based on the \textit{ordering of choices} heuristic detailed in \textit{Autologic} (Tennant, 1992).

The \textit{basic rule filter} is a simple binary feature that returns $1$ if the selected rule is applicable given the current problem state and $-1$ otherwise. Applicability here means whether or not the given rule could possibly generate the conclusion. For example, $\land$-introduction is only applicable when the sought conclusion is of the form $\phi\land \psi$; it can be safely rejected in all other cases.

\subsubsection{Atomic accessibility score}

This feature is based on the \textit{atomic accessibility heuristic} and \textit{backgrounding} techniques detailed in chapters 8 and 13 respectively of \textit{Autologic} (Tennant, 1992).

For problem states with an atomic conclusion $A$, this feature returns $-1.0$ if $A$ does not occur as an accessible positive sub-formula of any available premise. If $A$ is accessible from one of the premises, this feature returns values between $0$ and $1$ in descending order of preference (higher preference gives higher value) detailed in the \textit{Backgrounding Strategy} section of \textit{Autologic} chapter 13.

If the conclusion sought for the current problem state is non-atomic (i.e. is a sentence with at least one connective), this feature simply returns $0$.

\subsubsection{Major complexity score}

This feature ranks selected major premises for elimination rules by relative complexity to other available premises. Lower complexity premises are given preference over higher complexity ones, as they generally require less work to decompose in subsequent sub-problems.\textsuperscript{*}

The complexity $c$ of a sentence $s$ is computed according to (10).

\begin{equation}
    c(s) =
    \begin{cases}
        0 \hspace{1.0em} \text{if $s$ is atomic or $\bot$} \\
        1 + \sum\limits_{i} C(s_i)\\
    \end{cases}
\end{equation}

where $s_i$ denotes the $i$'th child node of the sentence (e.g. $A$ would be the first child node of $A\rightarrow B$).

The complexity score feature $C$ of state $q$ and action $a$ is then computed according to (11).

\begin{equation}
    C(q,a) = 1 - \frac{c(a_m)}{\max\limits_{s\in q_p}{c(s)}}
\end{equation}

where $a_m$ denotes the selected major premise for action $a$ and $q_p$ denotes the set of premises available in state $q$.

\textsuperscript{*}{\small This actually isn't entirely true if one takes into account the different in complexity between logical connectives. Breaking down a conditional premise $A\rightarrow B$ is, for example, always more difficult than eliminating a conjunction $A\land B$.}

\subsubsection{Weighted major complexity score}

The weighted major complexity score is computed similarly to the major complexity score detailed in the previous section. The only change is to the complexity function, given in (12).

\begin{equation}
    c'(s) =\left\{
     \begin{array}{l l}
       0 & \text{if $s$ is atomic or $\bot$} \\
       1 + \sum\limits_{i} C(s_i) & \text{if the primary connective of $s$ is $\land$}  \\
       1 + 2\sum\limits_{i} C(s_i) & \text{if the primary connective of $s$ is $\lor$}  \\
       1 + 3\sum\limits_{i} C(s_i) & \text{if the primary connective of $s$ is $\neg$}  \\
       1 + 4\sum\limits_{i} C(s_i) & \text{if the primary connective of $s$ is $\rightarrow$}  \\
       1 + 5\sum\limits_{i} C(s_i) & \text{if $s$ is a quantified sentence, i.e. $\forall$ or $\exists$}  \\
     \end{array}
   \right.
\end{equation}

The weights are added to account for the complexity difference between different logical connectives/quantifiers. The weights themselves can be chosen arbitrarily as long as the order is preserved.

\subsubsection{Shortest path to goal}

This feature aims to exploit the problem-as-graph representation by finding the shortest path from the root node of the selected premise to the root node of the conclusion. The intuition here is that premises which are closer in the graph to the conclusion will generally be more likely to yield fruitful results when chosen as the major premise of an elimination rule.

Dijkstra's algorithm is used to compute the shortest path in the problem graph. The feature value $P$ for state $q$ and action $a$ is computed according to (13).

\begin{equation}
    P(q,a) = \frac{1}{1 + d(a_m,q_c)}
\end{equation}

where $d(x,y)$ returns the shortest distance between node $x$ and node $y$, $a_m$ denotes the graph node for the selected major premise, and $q_c$ denotes the graph node for the conclusion.

Note that $d$ will return $0$ if $x=y$ and $\infty$ if there exists no path from $x$ to $y$.

\section{Experimental Results}

\subsection{Problem sets}

The first problem set used for training was a set of 86,156 auto-generated problems randomly sampled from all possible permutations of three predicates, three individuals (variables), and all first-order connectives. Only problems which were solvable or refutable using the baseline solver were included in the problem set. Training on this problem set was largely unsuccessful due to (i) the large amount of redundancy between samples and (ii) the overall lack of complexity of the problem set. Every problem was solvable with $T<20$, greatly limiting the scope of the search space to which the Q model was exposed. Because Q-learning is naturally reliant on the breadth of the state/action space seen by the model, it shouldn't be very surprising that such limitations would hinder its learning capability.

The second problem set used was a much smaller but richer set of 154 hand crafted problems titled \textit{posquestions} (Tennant, 1992). On roughly 40/154 problems, the baseline finished with $T>100$; on a further $8$ of those problems, $T>1000$. Four of those most "difficult" problems were selected and withheld to serve as a test set.

\subsection{Results}

The results in \textit{Table 1} and \textit{Table 2} show the average $T$ values over the entire problem set using 3-fold cross validation. Averages for \textit{Train} phase were updated only during the three validation passes for each epoch. For all experiments, $\alpha=1.0\times10^{-4}$ and $\gamma=0.9$. Training converged after 3-4 epochs for all feature sets. The primitive features discussed in section 4.4.1 are included in all experiments.

\textit{Figure 2} shows the difference in performance between the Q model and the baseline for problems with $T>100$ in the validation set; only features $A$ and $C$ are enabled.

A key for the feature set identifiers is given below.

{\small
\begin{itemize}[label={}]
    \item A - Atomic accessibility score
    \item B - Major complexity score
    \item C - Major weighted complexity score
    \item D - Shortest path to goal
\end{itemize}}

\begin{figure}
    \centering
    \includegraphics[scale=0.68]{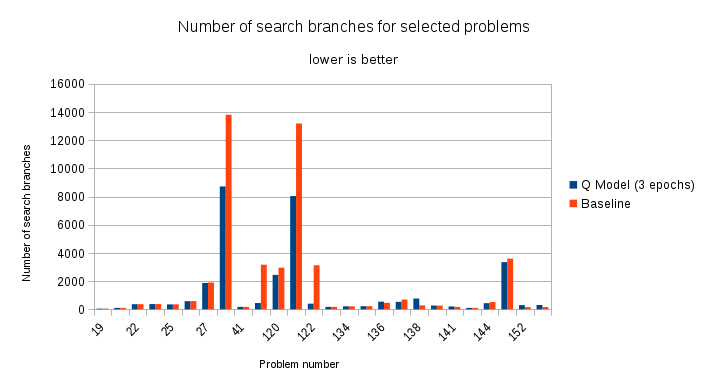}
    \caption{Validation $T$ values for selected problems}
    \label{fig:my_label}
\end{figure}

\begin{table}[H]
\centering
\caption{Average $T$ values after 3 epochs (features A,B,C,D)}
\label{t-vals-1}
\begin{tabular}{cccccc}
\multicolumn{1}{l}{} & \multicolumn{1}{l}{Baseline} & \begin{tabular}[c]{@{}c@{}}Q Model\\ A\end{tabular} & \begin{tabular}[c]{@{}c@{}}Q Model\\ B\end{tabular} & \begin{tabular}[c]{@{}c@{}}Q Model\\ C\end{tabular} & \begin{tabular}[c]{@{}c@{}}Q Model\\ D\end{tabular} \\
Train                & 375.3                        & 404.8                                               & 258.3                                               & \textbf{243.6}                                               & 292.5                                               \\
Test                 & 9503.1                       & 9503.1                                              & 4054.4                                              & \textbf{3336.4}                                              & 4010.9                                             
\end{tabular}
\end{table}

\begin{table}[H]
\centering
\caption{Average $T$ values after 3 epochs (combined features)}
\label{t-vals-2}
\begin{tabular}{cccccc}
\multicolumn{1}{l}{} & \multicolumn{1}{l}{Baseline} & \begin{tabular}[c]{@{}c@{}}Q Model\\ A,C\end{tabular} & \begin{tabular}[c]{@{}c@{}}Q Model\\ A,D\end{tabular} & \begin{tabular}[c]{@{}c@{}}Q Model\\ C,D\end{tabular} & \begin{tabular}[c]{@{}c@{}}Q Model\\ A,C,D\end{tabular} \\
Train                & 375.3                        & 270.2                                                 & 321.0                                                 & 4211.6                                                & 4210.8                                                  \\
Test                 & 9503.1                       & 3337.9                                                & 4010.9                                                & \textbf{185.9}                                                 & \textbf{185.9}                                                  
\end{tabular}
\end{table}

\begin{figure}
    \centering
    \includegraphics[scale=0.68]{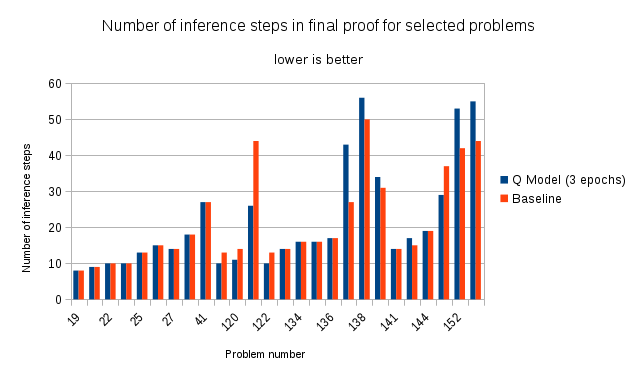}
    \caption{Validation $p$ values for selected problems}
    \label{fig:my_label}
\end{figure}

Isolated feature experiments in \textit{Table 1} show that the atomic accessibility feature ($A$) has surprisingly little to no effect on overall efficiency. However, both major complexity scores ($B$ and $C$), as well as the shortest path feature ($D$), show significant improvements. This seems to indicate that both features independently provide valuable information to the search algorithm.

The results in \textit{Table 2} are more difficult to interpret. While performance on the test set did improve dramatically with feature combinations $C,D$ and $A,C,D$, this came at the cost of significantly worse performance on the validation set. The best explanation for this is probably that the addition of multiple features caused the model to converge to a \textit{different local minimum} than the one found in the other experiments. It's entirely realistic to think that one policy might perform differently on a small subset of problems than it does on another. As is often the case with machine learning tasks, the best course of action here is likely to test the system on a larger set of problems.

\textit{Figure 3} shows the difference in \textit{quality} of proof between the Q model and the baseline. We consider "quality" in this context to be measured by the $p$ value, i.e. the \textit{length} of the proof. Shorter proofs are considered "better" than longer ones, as we generally want to find a proof in as few steps as possible.

Results here are also fairly mixed. In some cases, the Q model finds a better proof than the baseline; in other cases, the opposite is true. This makes it difficult to draw any conclusions about whether or not the learning algorithm helped improve proof search based on this metric.

\section{Conclusion and Future Work}

This work aimed to provide an exploratory study of applying reinforcement learning techniques to theorem proving in first order Core Logic (Tennant, 2017). One of the key limitations encountered was the limited amount of problems available for training. While automatic generation of problems for training is theoretically possible, in practice, it turned out to be quite difficult to do with reasonable time and space complexity. The primary problem set used was a set of only $154$ problems, which is likely not enough to derive any conclusive results about the efficacy of the learning algorithm. Future work should aim to solve this problem though either (i) improved problem generation (perhaps by randomly sampling syntax trees for sentences in a more efficient manner) or (ii) adapting an available problem set like TPTP (Sutcliffe, 2017) to Core Logic.

It would also be worthwhile to review the overall setup of the learning task for possible improvements. The reward function (9) was derived in order to satisfy the requirement of minimizing $p$ and $T$. It is not at all clear that it is an "optimal" reward function, nor even that there aren't metrics better suited to the task. This is, however, a well known problem in reinforcement learning; i.e. what is the best way to model the reward for an action such that the agent learns to maximize both \textit{long term} and \textit{short term} gain? This is a hard question to answer in the context of proof search and deserves further investigation.

Overall, this work does seem to show that applying reinforcement learning to proof search is a plausible approach. The linear Q model presented here, while fairly simplistic, was able to learn well enough to improve overall performance. Furhtermore, almost all experiments showed that the model was capable of generalizing techniques learned from simpler problems to those of higher difficulty in the test set. These are limited but nonetheless encouraging results, with much room for improvement and further exploration in the future.

\section{References}

Groenke, Brian. "AutoMoL". http://www.github.com/bgroenks96/AutoMoL. 2017-2018.

Howard, William A. (1980) [original paper manuscript from 1969], "The formulae-as-types notion of construction", in Seldin, Jonathan P.; Hindley, J. Roger, To H.B. Curry: Essays on Combinatory Logic, Lambda Calculus and Formalism, Boston, MA: Academic Press, pp. 479–490, ISBN 978-0-12-349050-6.

Kovács, Laura, and Andrei Voronkov. "First-order theorem proving and Vampire." International Conference on Computer Aided Verification. Springer, Berlin, Heidelberg, 2013.

Loos, Sarah, et al. "Deep network guided proof search." arXiv preprint arXiv:1701.06972 (2017).

McCune, William. "Prover9 and Mace4". http://www.cs.unm.edu/\texttildelow mccune/prover9/. 2005-2010.

McLaughlin, Sean, and Frank Pfenning. "Imogen: Focusing the polarized inverse method for intuitionistic propositional logic." International Conference on Logic for Programming Artificial Intelligence and Reasoning. Springer, Berlin, Heidelberg, 2008.

Schulz, Stephan. "System description: E 1.8." International Conference on Logic for Programming Artificial Intelligence and Reasoning. Springer, Berlin, Heidelberg, 2013.

Silver, David, et al. "Mastering the game of Go with deep neural networks and tree search." nature 529.7587 (2016): 484-489.

Sutcliffe, G. “The TPTP Problem Library for Automated Theorem Proving.” TPTP, University of Miami, www.cs.miami.edu/~tptp/. 2017.

Suttner, Christian, and Wolfgang Ertel. "Automatic acquisition of search guiding heuristics." International Conference on Automated Deduction. Springer, Berlin, Heidelberg, 1990.

Tennant, Neil. "Autologic." (1992).

Tennant, Neil. Core Logic. Oxford University Press. 2017.

Tennant, Neil. "The relevance of premises to conclusions of core proofs." The Review Of Symbolic Logic 8.4 (2015): 743-784.

Urban, Josef, Jiří Vyskočil, and Petr Štěpánek. "MaLeCoP machine learning connection prover." International Conference on Automated Reasoning with Analytic Tableaux and Related Methods. Springer, Berlin, Heidelberg, 2011.

Wang, Mingzhe, et al. "Premise Selection for Theorem Proving by Deep Graph Embedding." Advances in Neural Information Processing Systems. (2017).

Watkins, Christopher JCH, and Peter Dayan. "Q-learning." Machine learning 8.3-4 (1992): 279-292.

\end{document}